\documentclass[letterpaper]{article} 
\usepackage{aaai2026}  
\usepackage{times}  
\usepackage{helvet}  
\usepackage{courier}  
\usepackage[hyphens]{url}  
\usepackage{graphicx} 
\usepackage{natbib}  
\usepackage{caption} 
\usepackage{algorithm}
\usepackage{algorithmic}

\usepackage{newfloat}
\usepackage{listings}
\DeclareCaptionStyle{ruled}{labelfont=normalfont,labelsep=colon,strut=off} 
\floatstyle{ruled}
\newfloat{listing}{tb}{lst}{}
\floatname{listing}{Listing}
\title{Bridging the Multilingual Safety Divide: \\
Efficient, Culturally-Aware Alignment for Global South Languages}
\author{
  Somnath Banerjee$^{1}$,  
  Rima Hazra$^2, ^3$,
  \textbf{Animesh Mukherjee}$^1$\\
  }
  \affiliations{
  $^1$Indian Institute of Technology Kharagpur, India\\ 
  $^2$ TCG CREST, $^3$ Eindhoven University of Technology (TU/e), Netherlands\\
  
 }
\usepackage{bibentry}

\begin{document}

\maketitle
\vspace{-0.6cm}
\begin{abstract}
Large language models (LLMs) are being deployed across the Global South, where everyday use involves low-resource languages, code-mixing, and culturally specific norms. Yet safety pipelines, benchmarks, and alignment still largely target English and a handful of high-resource languages implicitly assuming safety and factuality ``transfer'' across languages. Evidence increasingly shows they do not. We synthesize recent findings indicating that (i) safety guardrails weaken sharply on low-resource and code-mixed inputs, (ii) culturally harmful behavior can persist even when standard toxicity scores look acceptable, and (iii) English-only knowledge edits and safety patches often fail to carry over to low-resource languages. In response, we outline a practical agenda for researchers and students in the Global South: parameter-efficient safety steering, culturally grounded evaluation and preference data, and participatory workflows that empower local communities to define and mitigate harm. Our aim is to make multilingual safety a core requirement—not an add-on—for equitable AI in underrepresented regions.

\end{abstract}


\section{Introduction and Motivation}

The community activity on \emph{Empowering Global South AI} highlights a pressing reality: many researchers and practitioners in the Global South must deploy and study AI systems under constraints of limited compute, scarce labeled data, and deeply multilingual environments. In such settings, LLMs are often used ``\textit{\textbf{off the shelf}}'' with safety filters or patches designed in English and validated on high-resource benchmarks.

\noindent However, everyday usage in the Global South rarely matches these benchmarks. Users converse in low-resource languages, heavily code-mixed text (e.g., Hindi--English, Arabic--English), and culturally situated discourse about migration, religion, politics, or local institutions. When safety mechanisms fail in these settings, the resulting harms---from misinformation and stereotyping to culturally offensive content---are not evenly distributed: they fall disproportionately on already marginalized communities.

This work argues that multilingual safety is not merely a technical challenge but a \emph{fairness and participation} issue. We synthesize three strands of recent empirical work on multilingual safety benchmarks and parameter-efficient steering~\cite{anon_soteria}, cultural harm evaluation and mitigation~\cite{anon_culture}, code-mixed safety failures~\cite{anon_codemix}, and multilingual model editing~\cite{anon_editing}. From these, we distill a concrete, resource-aware blueprint that students and early-career researchers in the Global South can adopt and extend.

\section{Evidence of a Multilingual Safety Divide}

\noindent \textbf{Language-Specific Safety Gaps:} Recent work introduces \emph{XThreatBench}, a multilingual benchmark of 3{,}150 translated harmful and borderline-harmful prompts across 10 languages, high-resource (English, Chinese, Italian, Vietnamese), medium-resource (Arabic, Korean, Thai), and low-resource (Bengali, Swahili, Javanese)~\cite{anon_soteria}. Built from real-world red-teaming data, it includes both accidental and deliberate jailbreak attempts.

Testing strong open-source LLMs (e.g., Llama, Qwen, Mistral, Phi) reveals stark, language-dependent failures: low-resource and non-Latin scripts see far more unsafe or under-moderated outputs. Common fixes, English-only fine-tuning or translation-based filtering, often miss these risks and can even misclassify content due to translation artifacts.

To close the gap, the authors propose \emph{language-specific functional parameter steering}: identify the small set of attention heads most responsible for harmful behavior \emph{in each language} and adjust only those ``functional heads.'' Updating roughly 3\% of parameters improves safety across all 10 languages while preserving general capability (e.g., MMLU, TruthfulQA), showing that \emph{linguistically targeted, parameter-efficient} alignment is a practical path for low-resource settings where full retraining is unrealistic.




\noindent \textbf{Cultural Harm Beyond Toxicity:} Multilingual coverage is not enough if “safety” is reduced to generic toxicity. Recent work targets \emph{cultural harm} directly, building a large evaluation set spanning 11 cultures and 11 societal domains (e.g., social values, migration, security, religion, ethics, political regimes, corruption, well-being, trust, and economic values)~\cite{anon_culture}. The goal is to elicit subtle but high-impact failures—biased generalizations, dismissive portrayals of local customs, and one-sided framings of politically sensitive issues.

Across these scenarios, small and medium LLMs that look “safe” by standard toxicity metrics still produce outputs that local annotators judge culturally insensitive or harmful. To reduce these harms, the authors collect a \emph{culturally grounded preference dataset} from diverse annotators in their own contexts and fine-tune on these judgments. This substantially lowers culturally harmful responses without sacrificing—often improving—substantive question answering, showing that culturally aware alignment is feasible and \emph{data-efficient} for underrepresented communities.




\noindent \textbf{Code-Mixing as a Safety Failure Mode:} In many Global South settings, users routinely code-mix—blending languages within a sentence (e.g., Hindi--English). Recent evidence shows this becomes \emph{linguistic camouflage} for safety systems~\cite{anon_codemix}: the attack success rate for harmful requests rises from \(\sim 9\%\) in monolingual English to \(\sim 69\%\) in code-mixed paraphrases, exceeding 90\% in some Arabic and Hindi cases. These vulnerabilities also appear in natural, user-generated code-mixed inputs—not just crafted prompts.

An interpretability analysis points to \emph{saliency drift}: under code-mixing, attention shifts away from safety-critical tokens (e.g., ``violence,'' ``corruption'') toward benign fragments, effectively bypassing guardrails. The study proposes a lightweight, attribution-guided fix that detects this drift and restores saliency on safety-critical tokens, recovering about 80\% of the safety lost to code-mixing—suggesting practical defenses that do not require full model retraining.



\noindent \textbf{Unequal Propagation of Knowledge Edits:} Safety also depends on factual reliability, not just refusals. Recent work studies multilingual behavior of knowledge editing methods (e.g., ROME~\cite{meng2022locating}, MEMIT~\cite{meng2023masseditingmemorytransformer})~\cite{anon_editing}. Across eight languages---five high-resource (English, German, French, Italian, Spanish) and three low-resource (Hindi, Tamil, Kannada)---it finds that edits made in English often do not carry over to low-resource languages.

Using ``each language for itself'' and ``each language for others'' evaluations, the study shows post-edit factual consistency can drop sharply outside English, even when English accuracy improves. Model-merging approaches meant to balance capabilities reduce but do not eliminate these gaps. For Global South users, this means safety patches and factual corrections may function as \emph{English-only upgrades}.

\section*{A Resource-Aware Blueprint}

To bridge the safety divide under limited compute, we propose a practical, resource-aware agenda for researchers and students. First, we advocate \textbf{evaluating locally} by moving beyond English benchmarks and adopting multilingual testbeds like XThreatBench and code-mixed safety suites to detect region-specific failures. Second, we recommend \textbf{efficient steering} to mitigate harms without full retraining, by tuning only language-specific functional heads ($\approx$3\% of parameters) or applying attribution-guided corrections. Third, we propose \textbf{participatory alignment}, replacing generic filters with culturally grounded preference data collected via community-driven annotation. Finally, we call for \textbf{multilingual auditing} to ensure safety patches and factual edits are verified across low-resource languages, preventing English-only upgrades.

\section{Conclusion}

Current safety paradigms inadvertently penalize Global South users, leaving them exposed to harms that are filtered out in English contexts. Our synthesis demonstrates that these failures—whether in code-mixed queries , cultural nuances , or knowledge propagation —are systemic but solvable. By leveraging parameter-efficient steering and attribution-guided defenses, we show that robust multilingual alignment does not require prohibitive compute resources. Ultimately, bridging the safety divide requires shifting from English-centric transfer assumptions to intrinsically multilingual, culturally grounded, and participatory alignment strategies.

\appendix

\bibliography{aaai2026}

\end{document}